\documentclass[10pt,twocolumn,letterpaper]{article}

\usepackage{cvpr}              
\usepackage{multirow}
\usepackage{adjustbox}
\usepackage{tabularx}
\usepackage{float}
\usepackage{makecell}
\newcommand{\gcell}[1]{\makecell{#1}}
\definecolor{cvprblue}{rgb}{0.21,0.49,0.74}
\usepackage[pagebackref,breaklinks,colorlinks,allcolors=cvprblue]{hyperref}
\usepackage{pifont}
\def\paperID{*****} 
\def\confName{CVPR}
\def\confYear{2026}

\title{Dual-branch Distilled Transformer for Efficient Asymmetric UAV Tracking}

\author{
Hongtao Yang\textsuperscript{\rm 1,\rm 2},
Bineng Zhong\textsuperscript{\rm 1,\rm 2}\thanks{Corresponding author.},
Qihua Liang\textsuperscript{\rm 1,\rm 2}\raisebox{-1.5pt}{*},
Yaozong Zheng\textsuperscript{\rm 1,\rm 2},\\
Xiantao Hu\textsuperscript{\rm 3},
Yuanliang Xue\textsuperscript{\rm 4},
Shuxiang Song\textsuperscript{\rm 1,\rm 2}\\
\textsuperscript{\rm 1}Key Lab of Education Blockchain and Intelligent Technology, Ministry of Education,\\
Guangxi Normal University, Guilin, 541004, China\\
\textsuperscript{\rm 2}Guangxi Key Lab of Multi-Source Information Mining and Security,\\
Guangxi Normal University, Guilin, 541004, China\\
\textsuperscript{\rm 3}Nanjing University of Science and Technology\\
\textsuperscript{\rm 4}Xi'an Research Institute of High Technology, Xi'an 710025, China\\
\texttt{\small yht@stu.gxnu.edu.cn, bnzhong@gxnu.edu.cn, qhliang@gxnu.edu.cn, yaozongzheng@stu.gxnu.edu.cn}\\
\texttt{\small xiantaohu@njust.edu.cn, xyl\_507@outlook.com, songshuxiang@mailbox.gxnu.edu.cn}
}

\begin{document}
\maketitle
\def\tracker{EATrack}
\begin{abstract}
Given the real-time demands of UAV tracking, many methods simplify the backbone to reduce computation, but this often weakens feature representation and degrades performance in complex scenarios. To alleviate this issue, we propose EATrack, an efficient and asymmetric UAV tracking framework centered around a teacher-guided dual-branch distillation strategy that enhances the feature expressiveness of the lightweight student model. Specifically, EATrack investigates two complementary perspectives of knowledge transfer: a spatially focused feature-level distillation that compensates for weakened representations by guiding the student to learn strong target representations, and a prediction-level distillation that enhances spatial localization by learning the teacher’s capability of accurate target localization. Furthermore, to enhance robustness against appearance variations, we introduce a fine-grained target-aware distillation strategy that selectively transfers the teacher’s target modeling capacity to the student. A temporal adaptation module is incorporated at inference to enhance robustness over time. Experiments on five UAV benchmarks demonstrate that EATrack achieves a favorable balance between accuracy and speed. 

\textbf{Code}---https://github.com/GXNU-ZhongLab/EATrack
\end{abstract}
    
\section{Introduction}
\label{sec:intro}

   \begin{figure}[t]
      \centering
      \includegraphics[height=7.5cm]{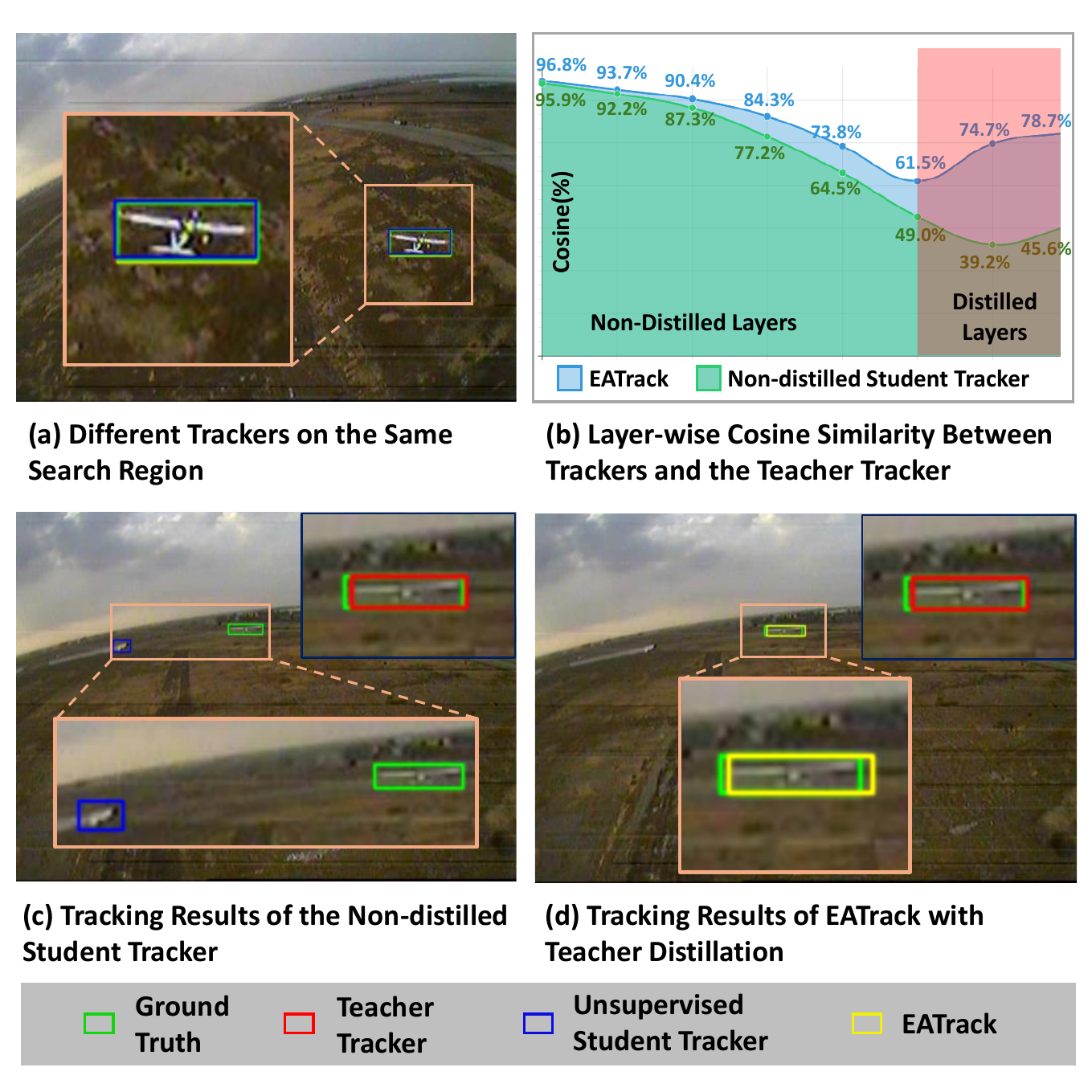}
       \caption{Impact of teacher tracker distillation. (a) Tracking results of different trackers on the same search image. (b) Layer-wise cosine similarity between trackers and the teacher tracker on the same search image. (c) The non-distilled student gradually loses the target. (d) Our tracker, benefiting from teacher distillation, achieves results comparable to the teacher tracker.}
       \label{fig:one}
    \end{figure}
Unmanned aerial vehicle (UAV) tracking plays a crucial role in a variety of real-world scenarios, such as wildlife protection \cite{wildlife}, environmental monitoring \cite{environmental_monitoring}, and traffic patrolling \cite{traffic}. Compared to general visual tracking, UAV-based tracking encounters more complex and dynamic conditions, including rapid motion, frequent occlusion, varying viewpoints, and small object sizes. Moreover, UAV platforms are often equipped with limited onboard computing resources, which place strict constraints on both model size and inference latency. These intertwined challenges highlight the urgent need for tracking algorithms that are both lightweight and robust, capable of maintaining reliable performance in demanding real-time aerial applications.

Existing UAV tracking methods have explored various strategies to balance speed and accuracy. Early approaches \cite{dcf_1, dcf_2, dcf_3} were based on discriminative correlation filters (DCFs), which offered high computational efficiency and were well-suited for onboard deployment. However, the limited modeling capacity and shallow representation of DCF-based methods made them vulnerable to appearance variations and occlusions. With the advent of deep learning, convolutional neural networks (CNNs) were introduced to enhance feature representation. These methods \cite{siamfc, hu2023transformer,hu2024toward, tctrack++} significantly improved robustness and generalization, but their local receptive field and fixed hierarchical structure constrained their ability to capture long-range dependencies. More recently, Vision Transformers (ViTs) have shown great promise in visual tracking due to their global modeling capability and rich semantic representation. Nevertheless, their high computational cost remains a major challenge for real-time UAV applications, prompting increasing interest in more lightweight and efficient Transformer designs. For instance, Aba-ViTrack \cite{Aba-ViTrack} reduces inference time through adaptive and background-aware token dropping, while AVTrack \cite{avtrack} selectively activates network components based on input content to enable adaptive computation. SGLATrack \cite{sglatrack} further streamlines ViTs by pruning redundant blocks via intra-model representational similarity. 
Although structural simplification improves computational efficiency, it often weakens the discriminative capacity of target representations, leading to degraded performance in complex or dynamic scenarios.

As shown in Fig.\ref{fig:one}, we conduct a controlled experiment to evaluate the effect of teacher-guided distillation on tracking performance. Both the distilled and non-distilled student trackers adopt the same lightweight architecture, differing only in whether teacher supervision is applied to the backbone layers. All trackers perform well in simple scenarios. However, the feature similarity analysis in Fig. 1(b) shows that distillation markedly enhances the alignment with the teacher’s features, particularly in the distilled layers. This strengthened feature consistency translates into improved robustness under challenging conditions: the distilled tracker maintains stable target localization comparable to the teacher, whereas the non-distilled student gradually drifts away from the target.

This observation underscores the importance of teacher-guided knowledge distillation in compensating for diminished feature representation caused by structural simplification. To address this, we propose a target-aware training framework that enhances feature expressiveness and localization precision of lightweight ViT-based trackers without adding inference overhead. Our framework integrates two complementary distillation strategies: spatially focused feature-level distillation transferring target-focused representations to guide fine-grained feature learning, and prediction-level distillation improving localization accuracy by aligning confidence distributions over target regions. Unlike ORTrack \cite{ortrack}, which focuses solely on feature-level distillation, our dual-branch design leverages the complementary benefits of both feature- and prediction-level guidance. Both strategies are applied exclusively during training, ensuring no additional computational cost during inference.
A temporal cue modeling mechanism is incorporated during inference to capture dynamic target variations and strengthen robustness in challenging scenarios. Unlike prior approaches relying solely on global guidance or overlooking region-specific learning, our method emphasizes fine-grained, target-aware distillation that concentrates supervision within the target region, effectively mitigating background interference. This dual-branch distillation design, combined with temporal adaptation, allows structurally pruned transformers to preserve competitive accuracy while maintaining architectural simplicity, clearly distinguishing our approach from existing distillation-based trackers.
Our contributions are summarized as follows:

\begin{itemize}
    \item A dual-branch distillation framework is specifically designed to selectively transfer target-specific knowledge from the teacher to the student model. By integrating spatially weighted feature-level supervision with masked prediction-level alignment, the framework effectively promotes richer feature representations and more accurate spatial localization within the student tracker.

    \item A fine-grained, target-aware training strategy is introduced to concentrate supervision within the target region, explicitly guiding the model to capture appearance variations. This focused supervision enhances robustness to subtle and dynamic changes, improving adaptability under complex motion and challenging visual conditions.

    \item An asymmetric backbone coupled with a temporal enhancement mechanism is designed to form a lightweight tracking framework that effectively captures target dynamics while maintaining a favorable balance between tracking accuracy and computational efficiency.
\end{itemize}

\section{Related Works}
\label{sec:formatting}

\subsection{UAV Tracking}
Existing UAV tracking methods can be roughly grouped into DCF-based, CNN-based, and Transformer-based approaches. DCF-based trackers \cite{kcf, autotrack} are known for their efficiency on low-power platforms but often fail to provide robust representations under complex scenarios. CNN-based Siamese trackers \cite{siamapn} enhance tracking robustness through deep feature extraction and pairwise similarity matching between the template and search region. However, their shallow backbones and rigid matching pipelines limit adaptability to aerial challenges with dynamic motion and appearance variations.
Transformer-based trackers~\cite{hu2025exploiting,hu2026curriculum,hu2025adaptive,shi2025mamba,yang2026motion,zeng2025explicit,li2026cadtrack} have recently drawn attention for their ability to capture global dependencies and support flexible interaction mechanisms. Methods such as OSTrack \cite{ostrack} and MixFormer \cite{mixformer} unify feature extraction and matching within a one-stream ViT architecture, achieving notable improvements in accuracy. However, the full ViT design incurs substantial computational cost, making it impractical for UAV platforms. To mitigate this, recent works pursue lightweight designs: Aba-ViTrack \cite{Aba-ViTrack} removes background tokens to reduce redundancy, AVTrack \cite{avtrack} conditionally activates modules based on input complexity, and SGLATrack \cite{sglatrack} prunes redundant layers via inter-layer similarity. These efforts aim at structural simplification to improve runtime efficiency.

However, structural simplification often impairs the propagation of rich feature information across layers, resulting in weakened target representations and degraded performance in complex scenarios. Differing from purely architectural compression, our work introduces a teacher-guided knowledge distillation framework that explicitly transfers strong feature representations and precise localization ability from a full ViT teacher model to a compact student. This target-aware distillation enables the student to preserve robust target modeling capabilities under structural constraints, while incurring no additional inference cost. Our method thus bridges the gap between efficiency and accuracy more effectively than prior lightweight designs.

\begin{figure*}
    \centering
    \resizebox{\linewidth}{!}{ 
    \includegraphics[height=0.5cm]{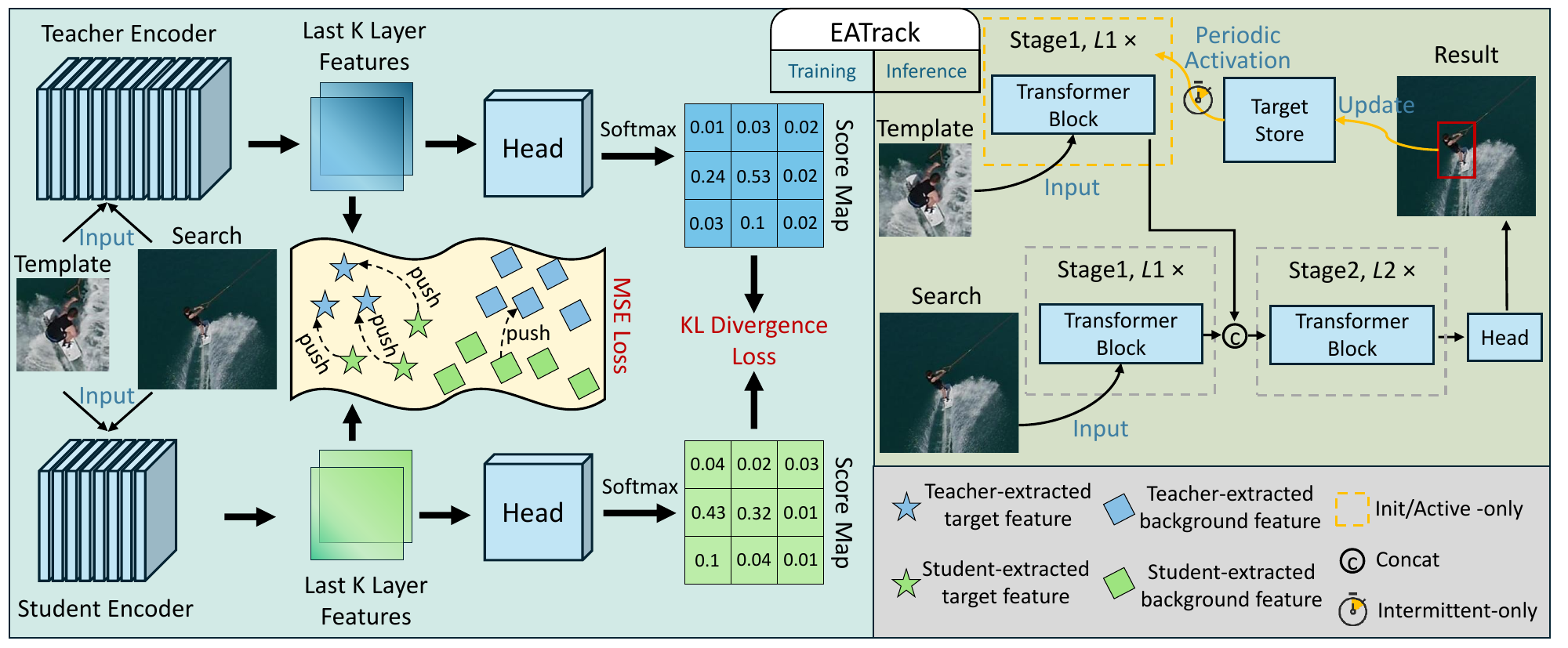}
    }
    \caption{Framework of the proposed EATrack. The left part shows our novel teacher-guided training strategy, where an \(L\)-layer teacher model provides effective feature-level distillation and prediction-level distillation to improve the target representation and localization of the \(L'\)-layer lightweight student. During inference (right), the student performs tracking independently in an asymmetric manner, with periodic template updates enabling temporal adaptability at ultra low cost. Notably, the target-aware training strategy is used only during training and incurs no extra computation overhead during inference.}
    \label{fig:two}
\end{figure*}

\subsection{Knowledge Distillation}
Knowledge Distillation (KD) was initially proposed for image classification, where a compact student network learns from the soft outputs of a larger teacher model \cite{hinton2015distilling}. Subsequent works extended KD to intermediate features, exploring strategies such as attention alignment \cite{zagoruyko2016paying}, relational knowledge transfer \cite{romero2014fitnets}, and multi-scale representation matching \cite{heo2019comprehensive}, achieving promising results in object detection and semantic segmentation. In the context of UAV tracking, ORTrack \cite{ortrack} introduces an Adaptive Feature-based Knowledge Distillation (AFKD) mechanism that modulates distillation strength according to sample difficulty. However, this adaptive strategy primarily operates at a global level and overlooks spatial priors inherently encoded in the teacher model. If such spatial cues are not explicitly leveraged, the student may lose focus on the target or suffer degraded localization under structural simplification, highlighting the need for target-aware guidance.

Our approach addresses this limitation by emphasizing a target-aware distillation paradigm that combines spatially weighted feature-level supervision with prediction-level guidance to explicitly transfer the teacher’s localization ability. Specifically, a fine-grained training strategy concentrates supervision within the target region, mitigating background interference and enhancing the model’s robustness against appearance variations. By integrating both feature- and prediction-level guidance, our framework preserves strong target modeling capabilities under structural constraints, bridging the gap between efficiency and accuracy more effectively than prior lightweight designs.

\section{Methodology}
This section presents a target-aware training strategy that enhances both representation and localization capabilities for the lightweight tracker. It introduces two complementary distillation branches at different levels, with the full-scale teacher used only during training. We also describe the student’s inference pipeline and the integrated temporal enhancement mechanism. As illustrated in Fig.\ref{fig:two}, the overall framework covers both training and inference stages.


\subsection{Preliminaries}
The inputs of our method include template image $\mathbf{Z} \in \mathbb{R}^{3 \times H_z \times W_z}$ and search image $\mathbf{S} \in \mathbb{R}^{3 \times H_s \times W_s}$, each independently processed by a shared patch embedding module to produce token sequences $\mathbf{X}_z \in \mathbb{R}^{N_z \times D}$ and $\mathbf{X}_s \in \mathbb{R}^{N_s \times D}$, corresponding to the template and search regions respectively.
The teacher model employs a full-scale vision transformer architecture with $L$ layers, which provides rich feature representations and precise localization capabilities. To achieve computational efficiency, the student model adopts a reduced depth of \(L' < L\) layers while retaining the same overall structure and tokenization process. This layer-wise correspondence facilitates the transfer of knowledge between teacher and student during training.

\subsection{Teacher-Guided Target-Aware Training}
To enhance the representational capacity and localization precision of the lightweight student model, a teacher-guided, fine-grained target-aware training strategy is introduced. This strategy focuses supervision within the foreground region, leveraging teacher predictions and structural cues to assign training signals to the target area. Such selective supervision helps the student recover discriminative representations often weakened by structural simplification, mitigating background interference. Based on this, a dual-branch distillation framework is constructed, consisting of feature-level and prediction-level distillation. These branches are complementary: the feature-level branch transfers fine-grained structural and appearance information, strengthening the student’s representational expressiveness; the prediction-level branch aligns confidence distributions and localization behaviors, refining spatial precision. Together, they facilitate a balanced transfer of detailed representation and positional accuracy, enabling the student to retain the teacher’s discriminative and localization capabilities while remaining structurally compact.

\textbf{Feature-level Distillation.}
In dense tracking scenarios, background noise can easily overwhelm the capacity of a lightweight student model, impairing its ability to discriminate the target. To address this, we guide the student to focus on target-relevant regions by transferring detailed feature representations from the teacher model in a spatially weighted manner. Specifically, a spatial weight mask $M_i$ is computed for each training sample based on the ground-truth bounding box, which activates the target region while suppressing the background. This mask is applied to both teacher and student features, enforcing the distillation loss to concentrate on informative foreground tokens and ignore distracting background cues. The constraint is formulated as a spatially weighted Mean Squared Error (MSE) loss:

\begin{equation}
    \begin{split}
\mathcal{L}_{\text{feat}} = \frac{1}{KB} \sum_{k=1}^{K} \sum_{i=1}^{B} \left\| M_i \cdot x_i^k - M_i \cdot y_i^k \right\|_2^2,
   \end{split}
\end{equation}

where $\mathbf{x}_i$ and $\mathbf{y}_i$ denote the feature representations of the student and teacher, $K$ is the number of distillation layers, and $B$ is the batch size. This selective alignment encourages the student to restore compact yet discriminative features degraded by pruning, thereby strengthening fine-grained target expressiveness.

\textbf{Prediction-level Distillation.}
While feature-level supervision transfers representational fidelity, accurate tracking further requires precise localization. To this end, we further supervise the prediction outputs by aligning the confidence distributions of the student and teacher within the target region using a masked Kullback–Leibler (KL) divergence loss. First, predicted confidence maps are converted into probability distributions via temperature-scaled softmax. Then, using the same binary mask derived from the ground-truth bounding box, the KL divergence is computed only over the target region:

\begin{equation}
    \begin{split}
\mathcal{L}_{\text{pred}} = \frac{1}{B} \sum_{i=1}^{B} \frac{\sum_{j} m_{i,j} \cdot \mathrm{KL}(p_{s,i,j} \| p_{t,i,j})}{\sum_{j} m_{i,j}},
   \end{split}
\end{equation}

where $p_{s,i,j}$ and $p_{t,i,j}$ represent the predicted probabilities at position $j$ from the student and teacher, respectively, and $m_{i,j} \in \{0,1\}$ is the target mask. This formulation ensures that the student focuses on accurately replicating the teacher’s localization behavior in regions that truly matter.

\textbf{Joint Effect and Training Efficiency.}
Together, the two distillation strategies form a complementary learning mechanism. Feature-level distillation strengthens the representational foundation of the student by preserving fine-grained structural details transferred from the teacher, thereby enhancing the ability to distinguish the target from cluttered backgrounds. Prediction-level distillation, on the other hand, focuses on aligning the spatial distribution of confidence scores, guiding the student to capture accurate target positions and boundary consistency. By jointly optimizing these objectives, the framework achieves a balanced transfer between detailed representation and spatial precision, allowing the student to preserve structural coherence and localization accuracy even under compact architectural constraints. Importantly, both distillation processes are applied exclusively during training and are discarded during inference, incurring no additional computational cost at test time. This target-aware distillation framework enables the student model to selectively inherit the teacher’s strengths by focusing on target regions, thereby learning robust representational clarity and precise localization rather than simply imitating the teacher’s behavior.

\subsection{Adaptive Asymmetric Inference}
During inference, our tracker adopts an asymmetric paradigm: the template and search frames are independently processed by the Stage 1 feature extractor. The extracted features are then concatenated and passed to Stage 2 for joint feature modeling, with the resulting fused representation fed into the prediction head to generate the final tracking outputs.
To maintain temporal awareness and adapt to appearance variations, a Target Store is introduced to accumulate reliable target representations over time. When the predicted bounding box surpasses a confidence threshold, the corresponding target feature is stored. At fixed intervals, representative embeddings are sampled from the Target Store and integrated with the original template representation. This dynamic update occurs through periodic activations, allowing the model to continuously refresh its understanding of the evolving target while retaining the computational efficiency of the asymmetric design.

Notably, Stage 1 template encoding runs only at initialization and periodic updates, ensuring effective temporal adaptation while balancing cost and tracking robustness.

\subsection{Training Loss}
To supervise the training process, we adopt a multi-term loss that combines standard tracking objectives with auxiliary distillation components. Specifically, the classification task is supervised using the focal loss \cite{lin2017focal}, while bounding box regression is optimized with a combination of $L_1$ loss and the GIoU loss \cite{rezatofighi2019generalized}. To further improve feature representation and localization precision in the student model, we incorporate two additional distillation terms that transfer knowledge from the teacher model at both the feature and prediction levels. The overall training loss is defined as:
\begin{equation}
    \begin{split}
L = L_{\text{cls}} + \lambda_1 L_1 + \lambda_2 L_{\text{GIoU}} + \lambda_3 L_{\text{feat}} + \lambda_4 L_{\text{pred}},
   \end{split}
\end{equation}

where $L_{\text{feat}}$ denotes the spatially weighted MSE loss computed on backbone features, and $L_{\text{pred}}$ represents the KL divergence loss applied to the predicted confidence distributions. The coefficients $\lambda_1 = 5$, $\lambda_2 = 2$, $\lambda_3 = 1$, and $\lambda_4 = 1$ control the relative importance of each loss term.

\begin{table*}[t]
    \centering
    \resizebox{\textwidth}{!}{
    \begin{tabular}{c|l|c|cc|cc|cc|cc|cc|cc}
    \toprule
     \multicolumn{1}{c|}{\multirow{2}{*}{Type}} & \multicolumn{1}{c|}{\multirow{2}{*}{Method}} 
     & \multicolumn{1}{c|}{\multirow{2}{*}{Source}}
     & \multicolumn{2}{c|}{DTB70} & \multicolumn{2}{c|}{UAVDT} & \multicolumn{2}{c|}{VisDrone} & \multicolumn{2}{c}{UAV123} & \multicolumn{2}{c}{UAV123@10fps}
     & \multicolumn{2}{c}{Avg.}\\
     
      & & & Prec. & Succ. & Prec. & Succ. & Prec. & Succ. & Prec. & Succ. & Prec. & Succ. & Prec. & Succ. \\
      \midrule
      \multicolumn{1}{c|}{\multirow{8}{*}{\rotatebox{90}{DCF-based}}} 
        & KCF \cite{kcf} & TAPMI 15 & 46.8 & 28.0 & 57.1 & 29.0 & 68.5 & 41.3 & 52.3 & 33.1 & 40.6 & 26.5 & 53.1 & 31.6 \\
        & fDSST \cite{fdsst} & TAPMI 17 & 53.4 & 35.7 & 66.6 & 38.3 & 69.8 & 51.0 & 58.3 & 40.5 & 51.6 & 37.9 & 60.0 & 40.7 \\
        & ECO\_HC \cite{eco_hc} & CVPR 17 & 63.5 & 44.8 & 69.4 & 41.6 & 80.8 & 58.1 & 71.0 & 49.6 & 64.0 & 46.8 & 69.7 & 48.2 \\
        & MCCT\_H \cite{mcct_h} & CVPR 18 & 60.4 & 40.5 & 66.8 & 40.2 & 80.3 & 56.7 & 65.9 & 45.7 & 59.6 & 43.4 & 66.6 & 45.3 \\
        & STRCF \cite{strcf} & CVPR 18 & 64.9 & 43.7 & 62.9 & 41.1 & 77.8 & 56.7 & 65.9 & 45.7 & 59.6 & 43.4 & 66.6 & 45.3 \\
        & ARCF \cite{arcf} & ICCV 19 & 69.4 & 47.2 & 72.0 & 45.8 & 79.7 & 58.4 & 67.1 & 46.8 & 66.6 & 47.3 & 71.0 & 47.1 \\
        & AutoTrack \cite{autotrack} & CVPR 20 & 71.6 & 47.8 & 71.8 & 45.0 & 78.8 & 57.3 & 68.9 & 47.2 & 67.1 & 47.7 & 60.0 & 40.7 \\
        & RACF \cite{racf} & PR 22 & 72.6 & 50.5 & 77.3 & 49.4 & 83.4 & 60.0 & 70.2 & 47.7 & 69.4 & 48.6 & 74.6 & 51.2 \\
        
      \midrule
      \multicolumn{1}{c|}{\multirow{9}{*}{\rotatebox{90}{CNN-based}}} 
      & HiFT \cite{hift} & ICCV 21 & 80.2 & 59.4 & 65.2 & 47.5 & 71.9 & 52.6 & 78.7 & 59.0 & 74.9 & 57.0 & 74.2 & 55.1 \\
      & SiamAPN \cite{siamapn} & ICRA 21 & 78.4 & 58.5 & 71.1 & 51.7 & 81.5 & 58.5 & 76.5 & 57.5 & 75.2 & 56.6 & 76.7 & 56.6 \\
      & P-SiamFC++ \cite{p-siamfc} & ICME 22 & 80.3 & 60.4 & 80.7 & 56.6 & 80.1 & 58.5 & 74.5 & 48.9 & 73.1 & 54.9 & 77.7 & 55.9 \\
      & TCTrack \cite{tctrack} & CVPR 22 & 81.1 & 61.9 & 69.1 & 50.4 & 77.6 & 57.7 & 77.3 & 60.4 & 75.1 & 58.8 & 76.0 & 57.8 \\ 
      & TCTrack++ \cite{tctrack++} & TPAMI 23 & 80.4 & 61.7 & 72.5 & 53.2 & 80.8 & 60.3 & 74.4 & 58.8 & 78.2 & 60.1& 77.3 & 58.8 \\
      & SGDViT \cite{sgdvit} & ICRA 23 & 78.5 & 60.4 & 65.7 & 48.0 & 72.1 & 52.1 & 75.4 & 57.5 & 86.3 & 66.1 & 75.6 & 56.8 \\
      & ABDNet \cite{abdnet} & RAL 23 & 76.8 & 59.6 & 75.5 & 55.3 & 75.0 & 57.2 & 79.3 & 60.7 & 77.3 & 59.1 & 76.7 & 59.1 \\    
      & PRL-Track \cite{PRL-Track} & IROS 24 & 79.5 & 60.6 & 73.1 & 53.5 & 72.6 & 53.8 & 79.1 & 59.3 & 74.1 & 58.6 & 75.2 & 57.2 \\

      \midrule
      \multicolumn{1}{c|}{\multirow{6}{*}{\rotatebox{90}{ViT-based}}}
      & LightFC \cite{litetrack} & KBS 24 & 82.5 & 63.9 & 81.6 & 59.3 & 79.7 & 61.4 & 84.2 & 65.9 & 83.1 & 64.0 & 82.2 & 62.9 \\
      & AVTrack-DeiT \cite{avtrack} & ICML 24 & 84.3 & 65.0 & 82.1 & 58.7 & \textbf{86.0} & \underline{65.3} & 84.8 & 66.8 & 83.2 & 65.8 & 84.1 & 64.4 \\
      & AVTrack-MD-DeiT \cite{avtrack-distill} & TCSVT 25 & 84.0 & 65.2 & \textbf{83.1} & \textbf{60.3} & 84.9 & 64.2 & 82.6 & 65.2 & 83.3 & 65.5 & 83.6 & 64.1 \\
      & ORTrack-D-DeiT \cite{ortrack} & CVPR 25 & 83.7 & 65.1 & 82.5 & 59.7 & 84.6 & 63.9 & 84.0 & 66.1 & 83.7 & 63.7 & 83.7 & 63.7 \\
      & SGLATrack-DeiT \cite{sglatrack} & CVPR 25 & 84.4 & 65.1 & 81.9 & 59.9 & 80.0 & 61.3 & 84.9 & \underline{66.9} & 82.6 & 65.5 & 82.8 & 63.7 \\
      
      & \gcell{\textbf{{{\tracker}}-ViT}} & \gcell{Ours} & \gcell{\textbf{85.3}} & \gcell{\underline{66.2}} & \gcell{81.8} & \gcell{60.0} & \gcell{\underline{85.6}} & \gcell{65.0}
      & \gcell{\underline{88.1}} & \gcell{\underline{66.9}} & \gcell{\underline{84.8}} & \gcell{\underline{66.7}} & \gcell{\underline{85.1}} & \gcell{\underline{64.9}}\\
      
      & \gcell{\textbf{{{\tracker}}-DeiT}} & \gcell{Ours} & \gcell{\underline{85.2}} & \gcell{\textbf{66.5}} & \gcell{\underline{82.7}} & \gcell{\underline{60.2}} & \gcell{85.3} & \gcell{\textbf{65.6}}
      & \gcell{\textbf{89.0}} & \gcell{\textbf{68.1}} & \gcell{\textbf{85.4}} & \gcell{\textbf{67.4}} & \gcell{\textbf{85.5}} & \gcell{\textbf{65.6}}\\

    \bottomrule
    \end{tabular}
    }
\caption{Comparison of precision(Prec.), success rate(Succ.), and speed(FPS) between {\tracker} and light weight trackers on DTB70, UAVDT, VisDrone2018, UAV123, and UAV123@10fps.
    Best in \textbf{bold}, second best \underline{underlined}.}
    \label{tab:results}
\end{table*}

\section{Experiments}
\subsection{Implementation Details}
\textbf{Model Family.}
To evaluate our proposed approach, we construct three tracker variants using ViT-tiny \cite{vit-tiny}, and distilled DeiT-tiny \cite{distill-deit-tiny} as backbones, named EATrack-ViT, and EATrack-DeiT. The teacher model adopts a transformer with $L = 12$ layers, while the student model uses a lighter version with $L' = 8$ layers. The prediction head follows a standard center-based design \cite{ostrack, mixformer} and consists of four Conv-BN-ReLU layers. Both during training and inference, the template and search region are uniformly resized to 128$\times$128 and 256$\times$256, respectively.

\textbf{Training strategy.}
We train EATrack on a combination of GOT-10k \cite{got-10k}, LaSOT \cite{lasot}, COCO \cite{coco}, and TrackingNet \cite{trackingnet}. Standard data augmentation techniques, such as horizontal flipping and brightness jittering, are applied during training. Each training sample is a triplet consisting of two template frames and one search frame. Training is conducted on four NVIDIA A800 GPUs, with each GPU processing 32 triplets, resulting in a total batch size of 128. We adopt the AdamW optimizer with a weight decay of \(10^{-4}\). The learning rate is initialized to \(4 \times 10^{-5}\) for the backbone and \(4 \times 10^{-4}\) for other modules. The model is trained for 300 epochs, with the learning rate decayed by a factor of 10 after epoch 240.

\textbf{Inference.}
Following common practice \cite{ostrack, transt} in visual tracking, a Hanning window penalty is applied during inference to incorporate spatial priors. The template feature is updated periodically, and the Target Store is refreshed only when the confidence exceeds a predefined threshold.

\subsection{Results and Comparisons}
\textbf{Datasets.}
To comprehensively evaluate the effectiveness of our method, UAV tracking benchmarks are used for evaluation, including DTB70 \cite{dtb70}, UAVDT \cite{uavdt}, UAV123 \cite{uav123}, UAV123@10fps \cite{uav123}, and VisDrone2018 \cite{visdrone}. Specifically, DTB70 contains 70 UAV sequences with diverse challenges such as fast motion, in-plane rotation, and background clutter. UAVDT is designed for vehicle tracking under various weather conditions, altitudes, and camera views, offering over 80K frames with high-resolution annotations. UAV123 is a large-scale UAV tracking benchmark consisting of 123 sequences with more than 112K frames. UAV123@10fps is derived by temporally downsampling UAV123 from 30 FPS to 10 FPS, simulating low-frame-rate tracking scenarios. VisDrone2018 includes drone-captured videos across urban and rural environments, characterized by dense objects, frequent occlusions, and complex camera motion.

\begin{table}[h]
\centering
\normalsize
\resizebox{\linewidth}{!}{
\begin{tabular}{c|c|ccccc}
\toprule
\multirow{2}{*}{Method} & UAV123 & \multicolumn{3}{c}{PyTorch Speed (fps)} & Params. & FLOPs \\
 & Succ. & GPU & CPU & TX2 & (M) & (G)\\
\midrule
{\tracker} & 68.1 & 241.9 & 97.4 & 33.6 & 6.20 & 1.87 \\
SGLATrack & 66.9 & 222.8 & 87.6 & 29.2 & 5.81 & 1.68 \\
ORTrack & 66.4 & 211.8 & 80.3 & 26.7 & 7.97 & 2.39 \\
\bottomrule
\end{tabular}}
\caption{Comprehensive Comparison of FPS, FLOPs, and Parameters under Consistent Hardware Settings.}
\label{tab:speed}
\end{table}

\textbf{Performance evaluation.}
We compare the proposed {\tracker}-DeiT with existing state-of-the-art lightweight trackers across five widely used UAV tracking benchmarks, including DTB70, UAVDT, VisDrone2018, UAV123, and UAV123@10fps. As shown in Tab.\ref{tab:results}, {\tracker}-DeiT achieves the highest precision and success scores on all five datasets, establishing new SOTA results in both metrics. Notably, it obtains a substantial gain on UAV123 and UAV123@10fps, surpassing the previous best method Aba-ViTrack by 2.6\% and 1.9\% in precision, respectively. On VisDrone2018, despite its challenging scenes with dense targets and heavy occlusion, our method still achieves a leading success score of 65.6\%.
Interestingly, while most ViT-based trackers exhibit strong performance, {\tracker}-DeiT consistently ranks first, highlighting the effectiveness of our architectural design and efficient token aggregation. The outstanding performance across diverse UAV benchmarks confirms the robustness and generalization ability of our tracker under various aerial tracking scenarios.

\textbf{Speed evaluation.}
For fair comparison, we measure the inference speed of recent representative trackers under the same hardware setup. Specifically, all models are tested on a machine equipped with an Intel Xeon Gold 6342 CPU and an NVIDIA A100 GPU, as shown in Tab.\ref{tab:speed}. To further assess real-world applicability, EATrack is also deployed on the embedded NVIDIA Jetson TX2 platform, where it achieves a real-time speed of 33.6 FPS without relying on acceleration tools such as TensorRT. These results collectively demonstrate both the efficiency and practical deployability of our tracker across diverse hardware environments.

   \begin{figure}[h]
      \centering
      \includegraphics[height=6cm]{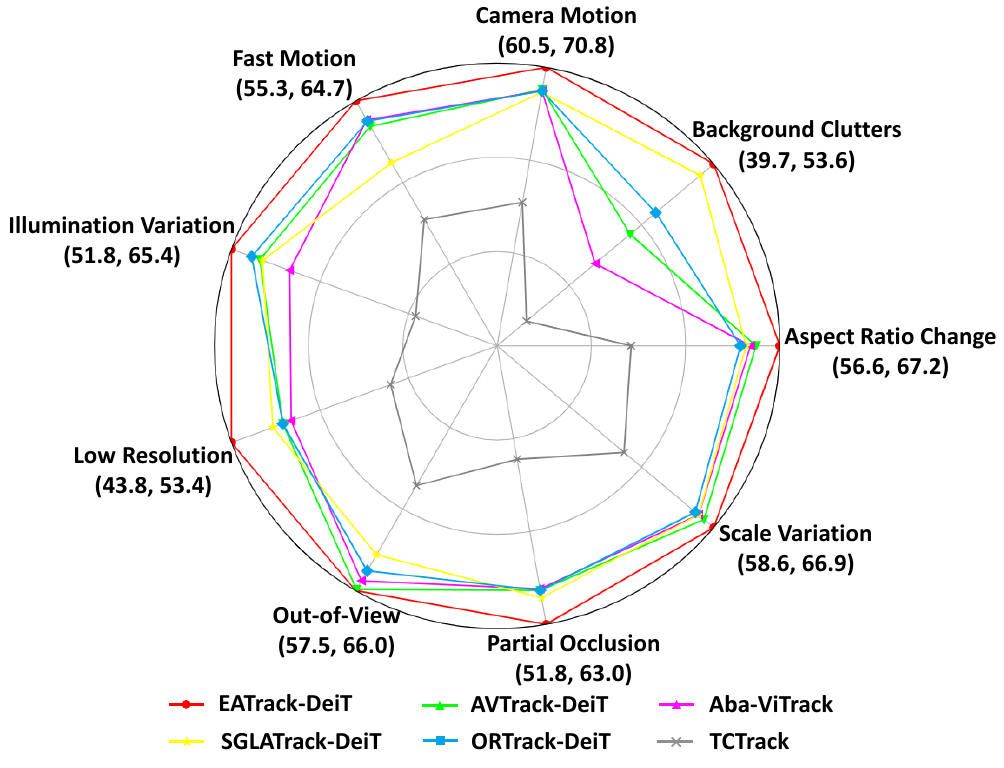}
       \caption{AUC scores of different attributes on UAV123.}
       \label{fig:leida}
    \end{figure}

\textbf{Attribute-Based Evaluation.}
To provide a more comprehensive and fair evaluation of our method, we compare {\tracker} against several recent state-of-the-art trackers on the widely used UAV123 dataset. As illustrated in the radar chart of Fig.\ref{fig:leida}, our method consistently demonstrates robust and balanced performance across all nine challenging attributes, outperforming existing trackers by an average AUC margin of 1.4\%. Notably, the most significant improvement is observed under low-resolution scenarios, where EATrack surpasses the second-best method, SGLATrack, by 2.5\%. Beyond that, EATrack also shows clear advantages under fast motion, camera motion, and illumination variation, indicating strong adaptability to highly dynamic and visually degraded conditions. Its stable and consistent performance across all attributes further suggests that the proposed teacher-guided distillation effectively enhances both representational robustness and target localization accuracy under diverse challenges.

\subsection{Ablation Study and Analysis}
To demonstrate the effectiveness of the proposed method, extensive ablation studies are conducted to evaluate the contribution and impact of each component.

\begin{table}[t]
  \centering
  \resizebox{\linewidth}{!}{
  \begin{tabular}{c|c|cc|cc}
    \toprule
    \multirow{2}{*}{\textbf{\#}} & \multirow{2}{*}{\textbf{Method}} & \textbf{Feature} & \textbf{Prediction} & \multicolumn{2}{c}{\textbf{UAV123}} \\
    & & \textbf{-level} & \textbf{-level} & \textbf{Prec.} & \textbf{Succ.} \\
    \midrule
    1 & \multirow{3}{*}{EATrack-DeiT}  & \ding{55} & \ding{55} & 86.9 & 66.1  \\
    2 & & \ding{51} &  & 87.6\(_{\uparrow 0.7}\) & 67.0\(_{\uparrow 0.9}\)  \\
    3 & & \ding{51} & \ding{51} & 89.0\(_{\uparrow 1.4}\) & 68.1\(_{\uparrow 1.1}\)\\
    \bottomrule
  \end{tabular}
  }
  \caption{Impact of Distillation on Model Performance.}
  \label{tab:ablation}
\end{table}

\textbf{Effectiveness of Knowledge Distillation.}
We conduct an ablation study on the UAV123 dataset to evaluate the effect of different knowledge distillation strategies, as shown in Tab.\ref{tab:ablation}. \#1 is a student tracker trained without any distillation. Introducing feature-level distillation (\#2) improves the expressiveness of target representations, leading to consistent gains in both precision and success. Further adding prediction-level distillation (\#3) brings an additional 1.4\% and 1.1\% improvement in precision and success, respectively. These two strategies are complementary and jointly contribute to a more robust student tracker.
In summary, the integration of complementary distillation strategies results in a well-balanced student model that achieves notable performance gains while maintaining inference efficiency, highlighting its potential for deployment in real-world, resource-constrained UAV scenarios.

\textbf{Impact of Layer Depth on Performance and Efficiency.}
To investigate the trade-off between the backbone depth and tracking performance, Tab.\ref{table:layer} compares student models with different backbone configurations under our distillation framework, where $L_1$ and $L_2$ denote the number of layers in Stage~1 and Stage~2, respectively.
We first analyze the impact of $L_1$. In configurations \#1–\#3, $L_2$ remains fixed while $L_1$ gradually increases. As the number of layers grows, performance consistently improves; however, the rate of improvement gradually slows down. Since each layer involves a fixed computational cost, the increases in parameters and FLOPs, as well as the reduction in speed, do not slow down correspondingly.
Next, we examine the effect of $L_2$. Configurations \#3–\#4 keep $L_1$ unchanged while varying $L_2$. It can be observed that a deeper joint feature modeling stage contributes more significantly to performance improvement. Nevertheless, in our design, Stage~2 entails higher computational complexity than Stage~1. Therefore, although \#3 and \#4 differ by only one additional layer, the efficiency degradation from \#4 to \#3 is much more pronounced compared to the gain from \#2 to \#3.
Based on these observations, configuration \#3 is adopted as the final backbone setting, achieving a favorable balance between accuracy and efficiency while maintaining the lightweight design objective of our framework.

\begin{table}[h]
\centering
\resizebox{\linewidth}{!}{
\begin{tabular}{c|*{2}{>{\centering\arraybackslash}p{1cm}|}ccccc} 
\toprule
\multirow{2}{*}{\#} & \multicolumn{2}{c|}{\textbf{Layerwise}} &  \multicolumn{2}{c}{\textbf{UAV123}} & \textbf{Speed} & \textbf{Params.} & \textbf{FLOPs}\\
 & \textbf{\textit{L}1} & \textbf{\textit{L}2} & Prec. & Succ. & (fps) & (M) & (G)\\
\midrule
1 & 4 & 2 & 87.5 & 66.4 & 280.6 & 5.31 & 1.59  \\
2 & 5 & 2 & 88.6 & 67.4 & 259.1 & 5.75 & 1.73  \\
3 & 6 & 2 & 89.0 & 68.1 & 241.9 & 6.20 & 1.87 \\
4 & 6 & 1 & 88.3 & 67.0 & 263.4 & 5.75 & 1.70  \\
\bottomrule
\end{tabular}
}
\caption{Impact of Model Depth on Performance, FPS, Parameters, and FLOPs.}
\label{table:layer}
\end{table}

\textbf{Impact of Temporal Cues on Cross-Dataset Tracking Performance.}
To thoroughly assess the contribution of temporal cues, comparative experiments are performed on four representative UAV tracking benchmarks, including DTB70, UAVDT, VisDrone, and UAV123. As summarized in Tab.\ref{tab:cues}, introducing temporal cues consistently brings noticeable gains across all datasets, particularly in dynamic or cluttered environments such as DTB70, UAVDT, and VisDrone. These scenarios typically involve fast camera motion, occlusion, and frequent appearance variations, where purely spatial modeling tends to be insufficient. The incorporation of temporal cues effectively alleviates these challenges by capturing motion continuity and stabilizing target localization. Notably, the improvements on DTB70 (+3.2\%) and VisDrone (+5.8\%) demonstrate that temporal modeling plays a pivotal role in enhancing robustness under complex motion patterns, while maintaining stable performance on relatively static datasets such as UAV123. Overall, these results confirm that temporal cues provide complementary information to spatial representations, leading to more reliable cross-dataset tracking performance.

\begin{table}[h]
  \centering
  \resizebox{\linewidth}{!}{
  \begin{tabular}{c|c|cccc}
    \toprule
    \multirow{2}{*}{\textbf{Method}} & \textbf{Temporal}  & \textbf{DTB70} & \textbf{UAVDT} & \textbf{VisDrone} & \textbf{UAV123}\\
     & \textbf{Cues}  & \textbf{Succ.} & \textbf{Succ.} & \textbf{Succ.} & \textbf{Succ.} \\
    \midrule
    \multirow{2}{*}{EATrack-DeiT}  & \ding{55} & 63.3 & 56.6 & 59.8 & 68.1  \\
      & \ding{51} & 66.5 & 60.2 & 65.6 & 68.1  \\
    \bottomrule
  \end{tabular}
  }
  \caption{Impact of Temporal Cues on Performance.}
  \label{tab:cues}
\end{table}




\textbf{Analysis and Evaluation of the Teacher Model.}
The teacher model follows the implementation of OSTrack, employing a distilled DeiT-tiny backbone. As shown in Tab.\ref{tab:one}, it achieves strong performance across the four major UAV benchmarks, benefiting from its full 12-layer feature modeling that provides rich target representation capability. Based on this teacher, a dual-branch knowledge distillation strategy is designed to transfer its representational ability to the student model. Considering the architectural asymmetry between the teacher and the pruned student, feature-level distillation is applied only to Stage 2, where joint modeling of template and search frames occurs, ensuring structural compatibility and effective feature alignment.
It is worth noting that the teacher model is lightweight and requires only moderate training cost, as it is trained once offline and does not participate in online inference. Therefore, the incorporation of the teacher supervision introduces negligible additional computational overhead and imposes no extra cost during tracking. Although the teacher model performs slightly below EATrack on certain challenging datasets, this performance gap mainly arises from the absence of a temporal adaptation mechanism, which in EATrack effectively improves robustness against appearance variations and motion disturbances during online inference.

\begin{table}[h]
  \centering
  \fontsize{9pt}{10pt}\selectfont
  \setlength{\tabcolsep}{1.9mm}
  \begin{tabular}{c|cccc|c}
    \toprule
    \multirow{2}{*}{\textbf{Method}} & \textbf{UAV123}  & \textbf{DTB70} & \textbf{UAVDT} & \textbf{VisDrone} & \textbf{Avg.}\\
     & \textbf{Succ.}  & \textbf{Succ.} & \textbf{Succ.} & \textbf{Succ.} & \textbf{Succ.}\\
    \midrule
    Teacher  & 68.3 & 63.7 & 56.0 & 64.1 & 63.0 \\
    \bottomrule
  \end{tabular}
  \caption{Performance Evaluation of the Teacher Model.}
  \label{tab:one}
\end{table}


  

\subsection{Qualitative Comparison and Visualization}

\begin{figure}[h]
      \centering
      \includegraphics[height=7cm]{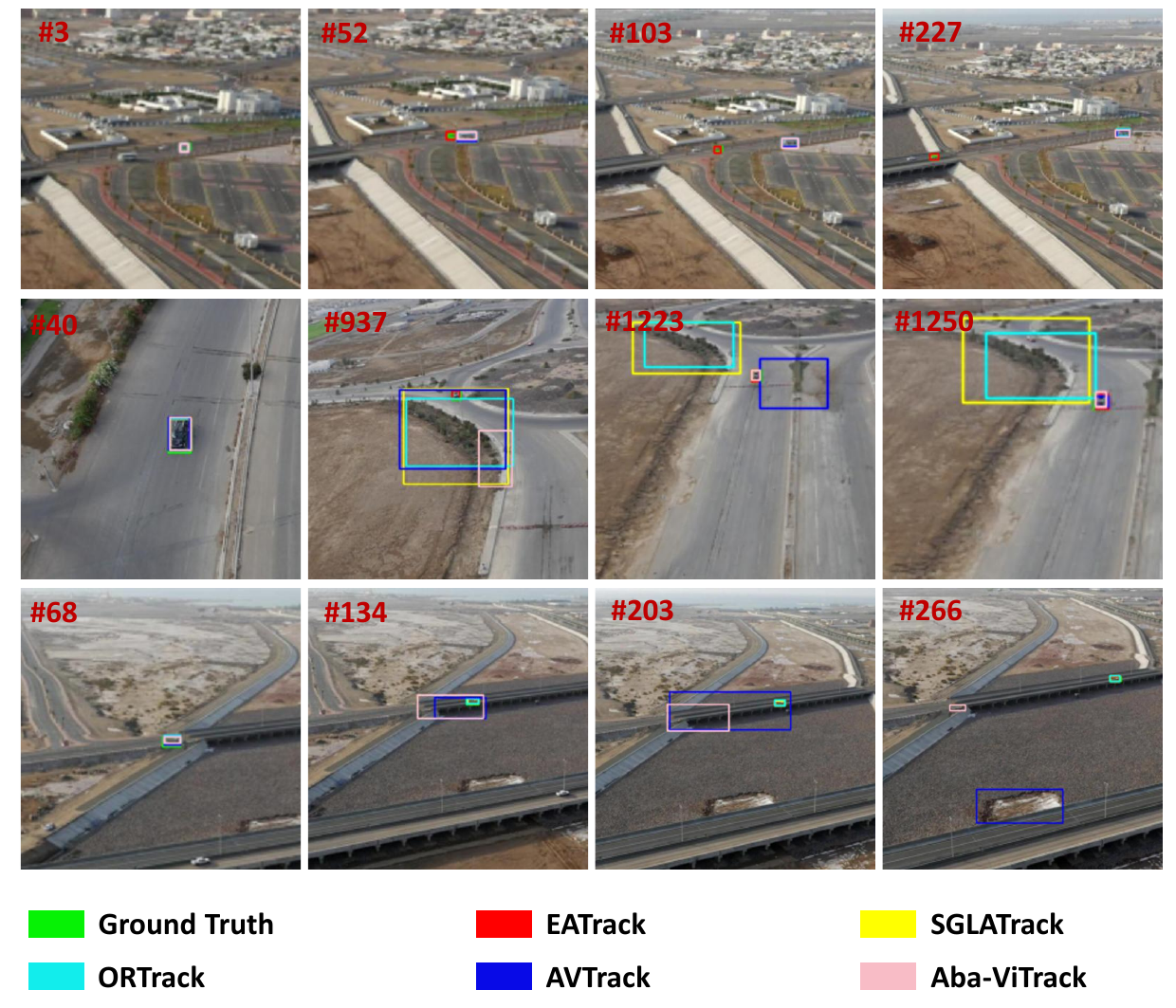}
       \caption{Qualitative comparisons of our tracker against other
 four SOTA trackers.}
       \label{fig:bbox}
    \end{figure}

\textbf{Qualitative results.}
To intuitively demonstrate the effectiveness of the proposed method, we present qualitative comparisons with four recent state-of-the-art trackers on the UAV123 dataset\cite{uav123}. As shown in Fig.\ref{fig:bbox}, our {\tracker} consistently exhibits robust tracking performance under various challenging scenarios. Notably, in scenes involving occlusion and interference from visually similar distractors, as illustrated in the first row, other trackers tend to lose the target and mistakenly track similar distracting objects. In contrast, our method maintains stable and accurate localization. These qualitative results provide strong empirical evidence supporting the effectiveness of our approach.
 
\begin{figure}[h]
      \centering
      \includegraphics[height=4.4cm]{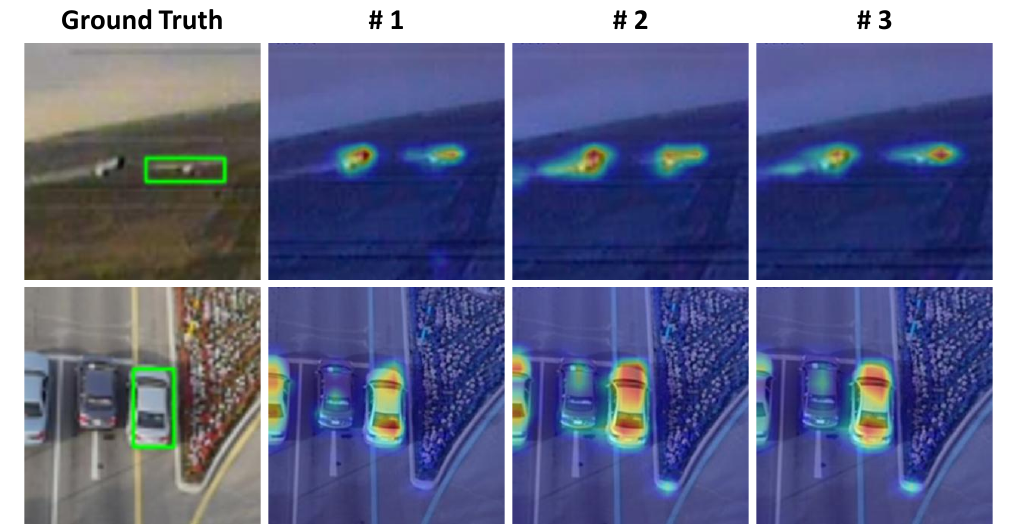}
       \caption{Comparison of Attention Maps from the Dual-Branch Distillation Architecture.}
       \label{fig:vis}
    \end{figure}

\textbf{Visualization of attention.}
To intuitively validate the effectiveness of the proposed complementary distillation strategies shown in Tab.\ref{tab:ablation}, we visualize the attention maps over the search region in Fig.\ref{fig:vis}, where \#1, \#2, and \#3 correspond to the configurations in Tab.\ref{tab:ablation}. It can be observed that the model progressively shifts its focus toward the target object from \#1 to \#3. In the presence of distractors with similar appearances, our distillation strategy enables the tracker to concentrate more accurately on the target, indicating enhanced robustness in target localization.

\section{Conclusion}
In this paper, we investigate how external guidance can enhance tracking under asymmetric architectures. We develop a teacher-guided training strategy that transfers feature- and prediction-level knowledge from an $L$-layer teacher to an $L'$-layer student, improving the student’s representation and localization without inference overhead. Feature-level distillation enhances spatial and temporal representations, while prediction-level distillation guides accurate and robust target localization; together, these complementary strategies reinforce both representation and prediction. Additionally, a periodic template activation scheme promotes temporal adaptability and efficiency. Based on these components, we construct a lightweight tracking framework called EATrack, where the student operates independently without teacher dependency. Experiments on multiple UAV benchmarks verify that EATrack achieves high accuracy while maintaining favorable inference speed.


{
    \small
    \bibliographystyle{ieeenat_fullname}
    \bibliography{main}
}


\end{document}